\documentclass[runningheads]{llncs}
\usepackage{graphicx}
\usepackage{amsmath,amssymb} % define this before the line numbering.
\usepackage{color}
\usepackage{pgfplotstable}
\usepackage{relsize}
\usepackage{xfrac}
\usepackage{diagbox}
\usepackage{sidecap}

\usepackage{xspace}

\usepackage{pgfplots}
\usepackage{pgfplotstable}
\usepackage{pgf,tikz}
\usetikzlibrary{arrows,shapes,decorations.markings}
\pgfplotsset{compat=1.9}

\newcommand{\extdata}[1]{\input{#1}}
\newcommand{\leg}[1]{\addlegendentry{#1}}

\newcommand{\correction}{correction\xspace}
\renewcommand{\paragraph}[1]{\noindent{\bf #1}\xspace}

\begin{document}
% \linenumbers
\pagestyle{headings}
\mainmatter

\title{Local Orthogonal-Group Testing} % Replace with your title

\titlerunning{Local Orthogonal-Group Testing}

\authorrunning{Ahmet Iscen, Ond{\v r}ej Chum}

\author{Ahmet Iscen \ \ \ \ Ond{\v r}ej Chum}
\institute{Visual Recognition Group, Faculty of EE, Czech Technical University in Prague}

\maketitle

\begin{abstract}
This work addresses approximate nearest neighbor search applied in the domain of large-scale image retrieval. Within the group testing framework we propose an efficient off-line construction of the search structures. The linear-time complexity orthogonal grouping increases the probability that at most one element from each group is matching to a given query. Non-maxima suppression with each group efficiently reduces the number of false positive results at no extra cost.
Unlike in other well-performing approaches, all processing is local, fast, and suitable to process data in batches and in parallel.
We experimentally show that the proposed method achieves search accuracy of the exhaustive search with significant reduction in the search complexity. The method can be naturally combined with existing embedding methods.

\end{abstract}

\newcommand{\head}[1]{{\smallskip\noindent\bf #1}}
\newcommand{\mypar}[1]{\noindent \textbf{#1}}
\newcommand{\ip}[2]{{#1}^{\top}{#2}}

\def \P    {\mathbf P}
\def \E    {\mathbb E}
\def \Var    {\mathbb V}
\def \real {\mathbb R}
\def \Xset {\mathcal X}
\def \Yset {\mathcal Y}
\def \Rset {\mathcal R}
\def \Sset {\mathcal S}
\def \Qset {\mathcal Q}
\def \x{\mathbf{x}}
\def \g{\mathbf{g}}
\def \w{\mathbf{w}}
\def \q{\mathbf{q}}
\def \b{\mathbf{b}}
\def \r{\mathbf{r}}
\def \m{\mathbf{m}}
\def \h{\mathbf{h}}
\def \y{\mathbf{y}}
\def \f{\mathbf{f}}
\def \z{\mathbf{z}}
\def \Y{\mathbf{Y}}
\def \A{\mathbf{A}}
\def \B{\mathbf{B}}
\def \C{\mathbf{C}}
\def \D{\mathbf{D}}
\def \G{\mathbf{G}}
\def \Q{\mathbf{Q}}
\def \H{\mathbf{H}}
\def \X{\mathbf{X}}
\def \S{\mathbf{S}}
\def \Z{\mathbf{Z}}
\def \V{\mathbf{V}}
\def \W{\mathbf{W}}
\def \N{\mathbf{N}}
\def \M{\mathbf{M}}
\def \K{\mathbf{K}}
\def \k{\mathbf{k}}

\def \cL {\mathcal L}

\def \one{\mathbf{1}}
\def \zero{\mathbf{0}}
\def \diag{\operatorname{diag}}

\def \sim{s}

\def \PX{\mathbf{P}_{\X}}
\def \PZ{\mathbf{P}_{\Z}}
\def \PZt{\mathbf{P}_{\Z^{\bot}}}
\def \PZW{\boldsymbol{\Pi}_{\Z,\W}}
\def \Real{\mathbb{R}}
\def \U{\mathbf{U}}
\def \u{\mathbf{u}}
\def \v{\mathbf{v}}
\def \s{\mathbf{s}}
\def \c{\mathbf{c}}
\def \1{\mathbf{1}}
\def \I{\mathbf{I}}
\def \a{\mathbf{a}}
\def \mub{\boldsymbol{\mu}}
\def \Ab{\bar{\mathbf{A}}}
\def \Bb{\bar{\mathbf{B}}}
\def \Cb{\bar{\mathbf{C}}}
\def \Db{\bar{\mathbf{D}}}
\def \alb{\boldsymbol{\alpha}}
\def \Pfn {\P_\text{fn}}
\def \Pfp {\P_\text{fp}}

\newcommand{\fix}{\marginpar{FIX}}
\newcommand{\new}{\marginpar{NEW}}

\def\sssp{\hspace{1pt}}
\def\ssp{\hspace{3pt}}
\def\msp{\hspace{5pt}}
\def\lsp{\hspace{7pt}}
\def\bsp{\hspace{12pt}}

\def \nbitsketch {P}
\def \nbitscalar {D}
\def \MaxIter {50}
\def \pinvc {\mathsf{pinv}}
\def \sumc {\mathsf{sum}}
\newcommand{\norm}[1]{\left\lVert#1\right\rVert}

\def\etal{\textrm{et al}.\,}
\def\ie{\emph{i.e.}~}
\def\vs{\emph{vs.}~}
\def\eg{\emph{e.g.}\xspace}

\def\defeq{\mathrel{\mathop:}=}

\newcommand*\rfrac[2]{{}^{#1}\!/_{#2}}

\newcommand{\ahmet}[1]{\textcolor{red}{#1}}
\newcommand{\ondra}[1]{\textcolor{green}{#1}}

\newcommand{\os}[1]{\textbf{#1}}
\newcommand{\ns}[1]{\textbf{\textcolor{red}{#1}}}

\newcommand{\T}{{\!\top}}
\newcommand{\mT}{{-\!\top}}

\def\roxf{$\mathcal{R}$Oxford\xspace}
\def\rox{$\mathcal{R}$Oxf\xspace}
\def\r1m{$\mathcal{R}$1M\xspace}

\section{Introduction}
\label{sec:intro}

In this paper, we are interested in approximate nearest neighbor search, specifically in large-scale image search. First, since the seminal paper of Sivic and Zisserman \cite{SZ03}, the image similarity was based on the bag-of-words approach\cite{PCISZ07,PCISZ08,JDS10a,JSHV10,GVSG10}. Efficient image retrieval was performed via inverted file structure~\cite{BL12}. Later, high dimensional non-sparse descriptors were introduced by VLAD~\cite{JDSP10} and Fisher vectors~\cite{PD07,PSM10}. Nowadays, image search is dominated by CNN descriptors\cite{BSCL14,TSJ15,GARL16,RTC16,RTC17}, which also use high dimensional non-sparse vectors to represent images. 
Image similarity is typically measured by cosine similarity of the descriptors, or equivalently by an Euclidean distance of $\ell_2$ normalized vectors. Efficient search in this case is performed by (approximate) nearest neighbor search. 

A number of methods exist for efficient search in high dimensional data, using a variety of approaches such as partitioning and embedding. In partitioning, the descriptor space is subdivided and only a small fraction of the data is actually considered for possible nearest neighbors, majority of the data is filtered out. These approaches include kd-tree and forests~\cite{ML14}, k-means tree~\cite{NS06}, LSH~\cite{IM98,GIM99,DIIM04}.
Embedding approaches find a mapping from the original descriptor space to some other (typically of lower dimension or binary) space, where the distance, or the ordering by the distance, can be very efficiently approximated. As an example of the embedding methods we mention LSH~\cite{IM98,GIM99,DIIM04} and other compact binary coding methods~\cite{LCL04,NPF12,WTF09,RL10}, product quantization~\cite{JDS11} and methods derived from it~\cite{GHKS13,KA14}. Commonly, a combination of partitioning and embedding, such as PQ-IVFADC~\cite{JDS11}, is adopted.
More recent works use neural networks to learn embeddings in a supervised manner~\cite{JZPG17,JZPG18}.

An alternative to these approaches are methods based on group testing.
Group testing was first used in World War II by the US army~\cite{D43}. Due to limited resources, the US army did not want to test each individual soldier for an STD; instead they combined several blood samples into a single mixture, and tested the mixtures. If the test was negative, the associated individuals were deemed sound. If the test was positive, then all the recruits that contributed to this mixture were tested individually. Since the percentage of infected soldiers was low, this procedure dramatically reduced the number of tests needed to screen the population of soldiers.

In  approximate nearest neighbor search, the individual blood samples are replaced with vectors, that are grouped into memory units. Each memory unit is represented by a vector, that is constructed from dataset vectors in the unit -- different means of construction can be used~\cite{IFGRJ14}, the simplest being a sum. In adaptive group testing~\cite{SFJ14,IFGRJ14,IAF16}, memory vectors are used for efficient pre-filtering of candidates. For the candidates, the exact similarity is then computed from the original vectors. To avoid storing the original vectors, methods of non-adaptive group testing were developed. Roughly, each input vector is stored in multiple memory units. The similarity of input vectors to the query is then estimated from a number of relevant memory units.

\begin{figure}
\begin{center}
\includegraphics[width=0.9\textwidth]{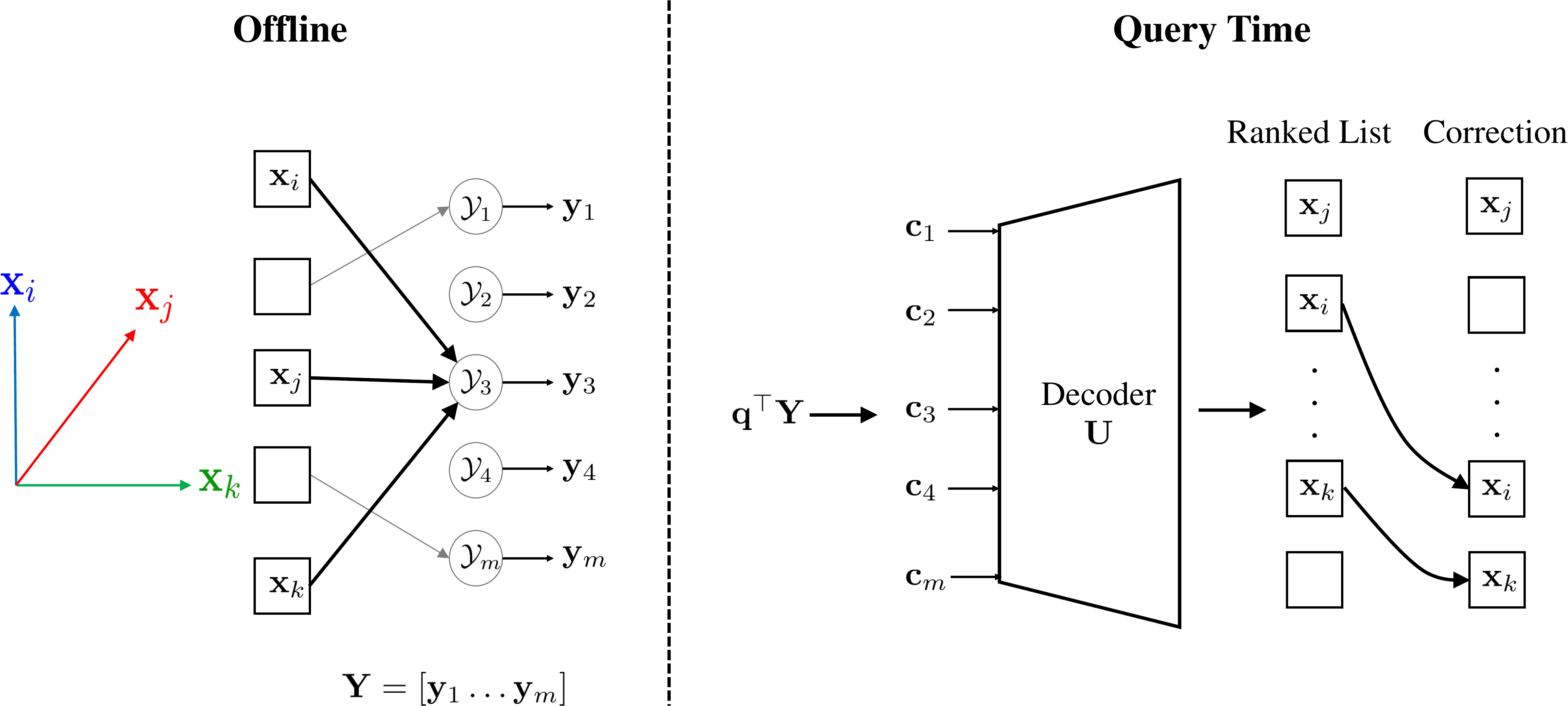}
\end{center}
\caption{The input data are represented by vectors $\mathbf{x}_1 \ldots \mathbf{x}_k$, By a greedy linear algorithm, vectors that are close to orthogonal are grouped into memory units $\mathcal{Y}_i$ represented by memory vectors $\mathbf{y}_i$. Sparse decoder matrix $\mathbf{U}$ captures the structure of the group assignment, and is also constructed in the off-line stage. At  query time, responses of the memory vectors to the query $q$ are evaluated and used to decode an estimate of the individual input vector similarities. A non-maxima suppression within each memory unit, called \correction, is applied to filter out possible false positives. In this particular example, $\x_i$ and $\x_k$ were suppressed as they share $\mathcal{Y}_3$ with higher ranked $\x_j$. 
\label{fig:intro}}
\end{figure}

In this paper, we are interested in non-adaptive group testing. The contribution of the paper is twofold: local processing of the dataset when constructing the search structure and novel grouping of the vectors that introduces additional constraints used in the scoring stage. The proposed method matches the quality of the dictionary learning methods~\cite{IRF16} that benefit from timely offline learning on the whole dataset, while preserving indexing efficiency of the basic group testing methods~\cite{SFJ14}.  The process of encoding and search is outlined in Fig.~\ref{fig:intro}.

\paragraph{Local processing.} The construction of the memory units and related decoding structures is linear in the size of the dataset. In order to encode the input data, the method (unlike \cite{IRF16}) does not need to see the whole dataset at once. This property makes it efficient in a streaming scenario, where the dataset is gradually increased in batches.

\paragraph{Vector grouping.} The group testing methods are efficient when the positive elements to be retrieved (syphilis infected individuals) are sparse in the dataset (population). By a simple and efficient greedy algorithm of selecting the vectors into groups, we try to minimize the chances that more than one element encoded in a single memory vector is positive. In the decoding stage, such a construction allows to suppress false positive responses. We call this process \emph{\correction}.

The rest of the paper is structured as follows. The problem is mathematically formulated and current approaches are reviewed in Section~\ref{sec:problem}. The proposed method is described in Section~\ref{sec:method}. Thorough experimental evaluation is presented in Section~\ref{sec:exp}.

\section{Problem statement}
\label{sec:problem}
In this section we detail the group testing setup and its applications in similarity search and image retrieval.

\subsection{Adaptive group testing}
\label{sec:adaptGT}

Group testing was introduced in similarity search by Shi \etal~\cite{SFJ14}. 
Assume that the dataset has $N$ $d$-dimensional vectors $\{\x_{i}\}_{i=1}^{N}$. 
The entire dataset is denoted by the $d\times N$ matrix $\X = [\x_{1},\ldots,\x_{N}]$.
Each vector $\x_i$ is $\ell_2$-normalized such that $\|\x_{i}\|=1$, $1\leq i\leq N$. 
The similarity measure between a given query $\q$ and a dataset vector $\x_i$ is computed by the scalar product $s_i = \x_{i}^{\top}\q$. 

The goal of group testing is to infer the similarities $s_i$ efficiently through $M$ ($M \ll N$) group measurements and a decoder.
It has three stages. 
The encoding stage first assigns vectors to groups.
Shi~\etal~\cite{SFJ14} define the encoding matrix $\G$ as a $M \times N$ matrix which keeps group assignments, such that $\G_{ji} = 1$ if $\x_i$ belongs to $j$th group.
$\G$ is populated such that each vector is randomly assigned to $m$ groups and each group has exactly $n$ vectors:
\begin{equation}
\label{eq:MmNn}
M = \frac{mN}{n}.
\end{equation}
Then the group vectors are created based on their assignments:
\begin{equation}
\label{eq:encoding}
\Y = \X \G^{\top}.
\end{equation}
This is equivalent to summing all image vectors assigned to a group:
\begin{equation}
\label{eq:sum}
\y_j = \sum_{\x\in\Yset_j} \x, 
\end{equation}
where $\Yset_j = \{\x_1,\dots,\x_n\}$ is the set of $n$ vectors from $\X$ assigned to $j$th group, for $j = 1,\dots,M$. 

The group measurement and decoding stages are performed during the query time.
For a given query, the $M$ group measurements are computed
\begin{equation}
\label{eq:measurements}
\c = \q^{\top}\Y,
\end{equation}
and pass them to a decoder to approximate $N$ image similarities:
\begin{equation}
\label{eq:decoding}
\hat{\s} = \c\U = \q^{\top}\Y \U.
\end{equation}

The authors consider $\U = \G^{\top}$ in their work with an extra back-propagation step. 
In summary, when a new query is given, they i) compute its similarities with group vectors~\eqref{eq:measurements},  ii) estimate the image vector similarities~\eqref{eq:decoding}, iii) perform a back propagation where the exact similarities with top ranked image vectors are computed, iv) rank the images according to their similarity.

Iscen~\etal~\cite{IFGRJ14} use a different setup but a similar idea. 
They assign each image vector to a single group (called memory units), thereby reducing the number of groups.
Furthermore, the properties of randomly assigned memory units are theoretically analyzed. 
Assume that memory unit $\Yset_j = \{\x_1,\dots,\x_n\}$ stores $n$ dataset vectors.
Query $\q$ is related to $\x_1$, such that $\q = \alpha\x_{1}+\beta \Z$ where $\alpha$ is the similarity between $\q$ and $\x_{1}$, and $\Z$ is a random vector orthogonal to $\x_{1}$ and $\|\Z\|=1$.
When sum aggregation~\eqref{eq:sum} is used to create a group representative vector (called \textit{memory vector}) $\y_j$ from $\Yset_j$, the inner product between $\q$ and $\y_j$ becomes
\begin{equation}
\label{eq:ipNaive}
\ip{\y_j}{\q} = \ip{\x_1}{\q} + \alpha \sum_{\x_i \in \Yset_j \backslash \x_1} \ip{\x_i}{\x_1} + \beta \sum_{\x_i \in \Yset_j \backslash \x_1} \ip{\x_i}{\Z},
\end{equation}
where $\Yset_j \backslash \x_1$ denotes all vectors in $\Yset_j$ except $\x_1$.

The main source of noise for~\eqref{eq:ipNaive} comes from the middle term, which is basically the interference between $\x_1$ and the other $\x_i$ in $\Yset_j$. 
In an attempt to eliminate this noise, a construction of the memory vectors by computing the pseudo-inverse of all vectors assigned to the group is proposed in ~\cite{IFGRJ14}:
\begin{equation}
%\boxed{
\y_j=([\x\in\Yset_j]^{+})^{\top}\1_{n},
%}
\label{eq:pinv}
\end{equation}
where ${+}$ denotes \textit{Moore-Penrose pseudo-inverse}~\cite{RM71}.
This construction is shown to perform better than the sum construction both theoretically and empirically under some mild conditions. At the final stage, they also re-rank all vectors of the highest scoring groups by computing their true similarity with the query vector.

Both methods indeed perform what is called \textit{adaptive group testing} composed of two steps. The first step computes similarities with group representatives. These indicate which vector similarities are worth being investigated. The second step is a verification process computing the true similarities for these candidate vectors.
Even though this strategy gives very accurate results with high efficiency, it requires group representatives as well as all database vectors to be kept in memory for the second adaptive step.

\subsection{Non-adaptive group testing}
\label{sec:nonAdaptGT}

A more modern view on group testing, called \textit{non-adaptive group testing}, was adopted in other computer science related fields thanks to the advancements made in compressed sensing~\cite{ABJ14,BBIAD11,CHKV09,GI10,SeJ10}.
This approach argues that the identification of the infected individuals is possible just from the results of the group tests realized in the first step. There is no need of a second verification step.

A more recent dictionary learning based approach~\cite{IRF16} applies non-adaptive group testing to similarity search.
Group testing is defined as an optimization problem, where group vectors $\Y$ and the decoder $\U$ are optimized jointly.
Unlike previous approaches~\cite{SFJ14}, there are no assignment or construction constraints. 
The only constraint is to have a sparse $\U$ to reduce the number of vector multiplications during the query time.  
The solution is found by a dictionary learning optimization algorithm which yields a sparse decoding matrix $\U$:
\begin{equation}
\label{eq:dictLearn}
\begin{aligned}
& \mathop{\text{minimize}}_{\Y,\U}
& & \frac{1}{2}\norm{\X - \Y\U }^2_F + \lambda \norm{\U}_1\\
& \text{subject to}
& & \norm{\y_{k}}_2 \leq 1 \text{ for all } 0 \leq k < M. \\
\end{aligned}
\end{equation}

This method has a good search efficiency in terms of complexity and memory footprint, without compromising the search accuracy. 
Original database vectors need not be stored in the memory anymore since there is no need for re-ranking with original similarities.
Nevertheless, its main weakness is its offline complexity.
The solution to~\eqref{eq:dictLearn} requires the entire dataset to be available, which means that it needs to be recomputed as new data becomes available.
Additionally, complexity of solving~\eqref{eq:dictLearn} grows dramatically as $N$ and $M$ increases, limiting its scalability for very large-scale scenarios.

\section{Our method}
\label{sec:method}
This section describes our contributions to group testing in this paper.
Our framework groups orthogonal vectors together, which allows to have an efficient correction step without computing the true similarities of dataset vectors. Both, the encoder and decoder are learned locally from a subset of the dataset.
We use the terminology introduced by~\cite{IFGRJ14} throughout the paper. Groups are called memory units, and group representatives are called memory vectors.

\subsection{Orthogonal memory units}
\label{sec:omv}

Random assignment~\cite{SFJ14} of the input vector to memory units is a basic assignment strategy: 
$m$ permutations of $\{1,\ldots,N\}$ denoted by $\{\boldsymbol{\pi_{k}}\}_{k=1}^{m}$ are drawn at random. 
For the $k$-th permutation, the vectors whose indices are $\pi_{k}((\ell-1)n+1),\ldots,\pi_{k}(\ell n)$, $\ell\in\{1,\ldots,N/n\}$, are grouped into one memory unit, assuming that $n$ divides $N$. 
Random assignment is a convenient choice for large-scale datasets or streaming data due to its low complexity and locality.
Other alternatives, such as assigning data based on k-means clustering~\cite{IFGRJ14} or kd-tree partitioning~\cite{IAF16} show that grouping similar vectors together improve efficiency for adaptive group testing.
However, these methods have a potentially expensive extra processing step and a verification step with re-ranking with true similarities is needed.

Our method heads the opposite direction. We propose to create a memory unit
so that it contains mutually orthogonal vectors.
The reason for such a construction is twofold, first minimizing the interference~\cite{IFGRJ14} between the vectors stored in a memory unit (\ref{eq:ipNaive}) and increasing the chances that there is only a single matching vector to a query in each memory unit. As we will show later, this property allows us to correct false positives and significantly improve the retrieval accuracy.

Instead of grouping $n$ random vectors (selected via random permutation) as in~\cite{SFJ14}, a random chunk of $k n$ vectors (again via random permutation) is selected. Within a chunk (considering only vectors from that chunk), $k$ memory units of $n$ vectors are constructed by a greedy approach.
The memory units are initialized with randomly selected vectors. In each iteration of the algorithm, each of the $k$ memory units is greedily extended by one vector, that is ''the most orthogonal'' to the vectors already assigned to that particular memory unit.
%~\footnote{Matlab source code provided in supplementary material} 

Due to the greedy nature of the algorithm, the assignment is not globally optimal. Obtaining the globally optimal assignment is intractable. Our experiments show that in practice the group assignments are sufficiently independent: sum~\cite{SFJ14} and pseudo-inverse memory vector construction for orthogonal memory units are equally good.

The time complexity of the orthogonal assignment is the same as the time complexity of the random assignment, and the algorithm is fast in practice. This makes the algorithm an efficient option for large-scale scenarios.
Since it works on small chunks of the dataset independently, it easily handles additional (streaming) data and is easily parallelized (unlike~\cite{IRF16}).

\subsection{Local decoder}
\label{sec:localDec}

After the assignment proposed in Section~\ref{sec:omv}, $M$ memory vectors are constructed with pseudo-inverse~\eqref{eq:pinv} and stored in a $d \times M$ matrix $\Y = [\y_1 , \dots, \y_M]$. During the query time, the goal is to approximate similarity of the query vector $\q$ and each individual input vector $\x_i$. This is achieved through a decoder matrix $\U$, so that
$$
\q^{\top} \X \approx \left(\q^{\top} \Y\right) \U\mbox{.}
$$

Shi~\etal~\cite{SFJ14}, propose multiple decoding schemes for a given query.
The first proposal is to take the pseudo-inverse of the sparse encoder $\G$ (see Section~\ref{sec:adaptGT}):
$\U = \G^{+}$.
This is a costly operation which involves computing and storing the dense pseudo-inverse of $\G$ in the query time.
Thus, they use a simpler sparse decoder
\begin{equation}
\U = \G^{\top} \label{eqn:GT}
\end{equation}
and an extra back-propagation step.

In the dictionary learning approach of Iscen~\etal~\cite{IRF16}, group vectors $\Y$ and decoding matrix $\U$ are estimated by a joint optimization~\eqref{eq:dictLearn}, which is extremely time demanding. Nevertheless, once the group vectors are constructed and fixed, the decoder matrix $\U$ is estimated by each column $\u_i$ independently by solving the system of linear equations
\begin{equation}
\label{eq:decoder}
\x_i = \Y \u_i.
\end{equation}
For efficiency reasons, it is important that $\u_i$ are sparse vectors. Let $\Sset(i)$ be the set of indices of the non-zero elements in $\u_i$. For a given $\Sset(i)$, the solution is found by solving a system of $||\Sset(i)||$ linear equations
\begin{equation}
\label{eq:localDecoder}
\x_i = \Y_{\Sset(i)} \u_{i,\Sset(i)} \mbox{,}
\end{equation}
where $\Y_{\Sset(i)}$ are the columns of $\Y$ whose indices are stored in $\Sset(i)$ and $\u_{i,\Sset(i)}$ is a vector composed of elements of $\u_i$ with indices $\Sset(i)$.

\begin{figure}
\begin{center}
\includegraphics[width=0.5\textwidth]{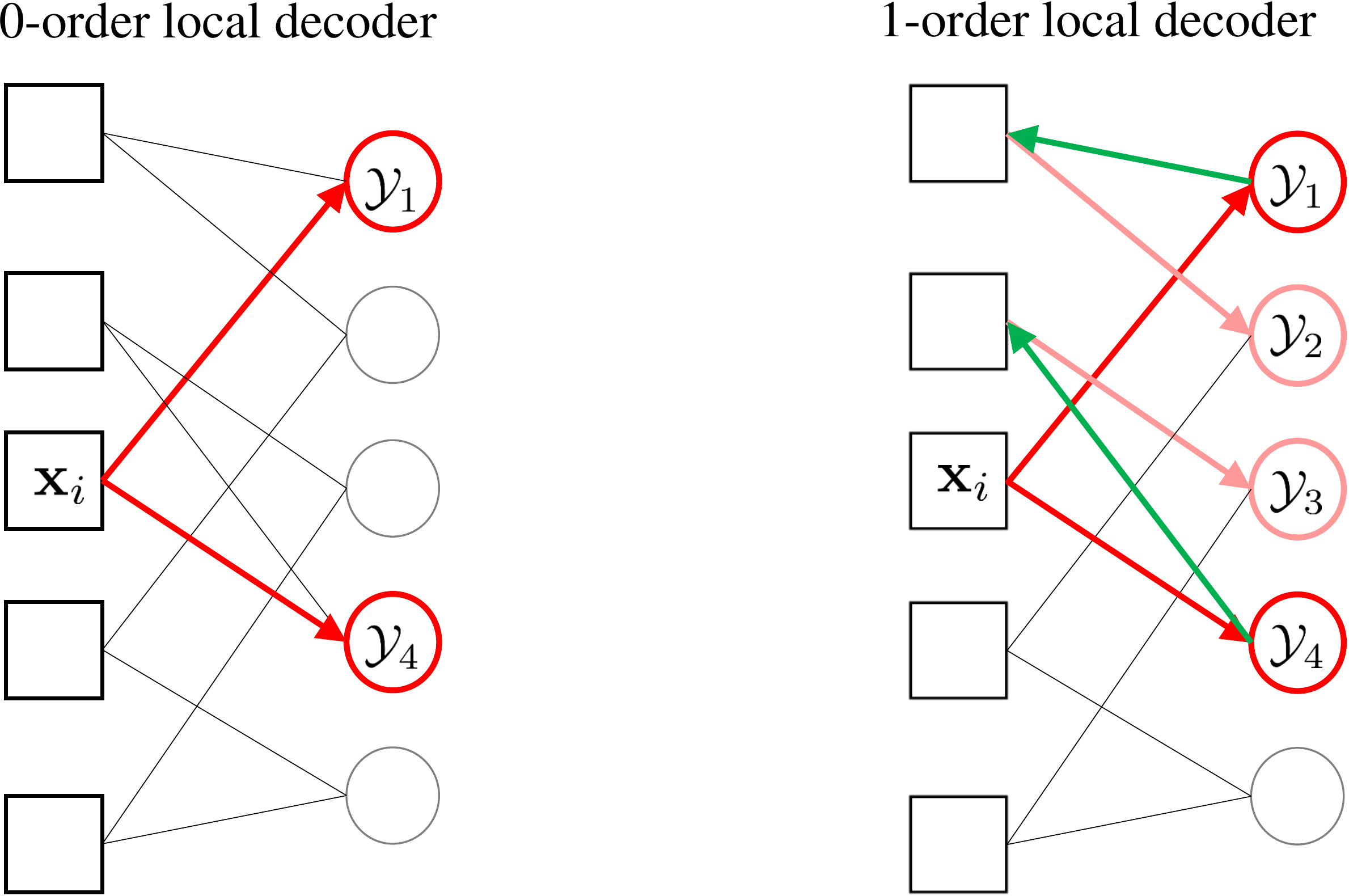}
\end{center}
\caption{\textbf{Left:} An illustration of 0-order local decoder for $\x_i$. $\Sset_0(i)$ contains indices of memory units that $\x_i$ is assigned to. \textbf{Right:} An illustration of 1-order local decoder for $\x_i$. $\Sset_1(i)$ contains indices of the memory units that contain any input vector $\x_j$ assigned to any memory unit with indices $\Sset_0(i)$. \label{fig:localDecoder}
}
\end{figure}

We propose to construct the set of indices $\Sset(i)$ based on the groupings of the input vectors into the memory units. Consider the bipartite graph $B$ illustrated in Fig.~\ref{fig:localDecoder}. In this graph, one type of nodes corresponds to the input vectors, the other type corresponds to memory units. There is an edge between two nodes if the corresponding input vector is a member of the memory unit. It is natural to attempt to decode input vector $\x_i$ with memory units that vector is assigned to. We call this 0-order local decoder. For this decoder, the $\Sset_0(i)$ contains indices of memory units connected by a single edge to $\x_i$ in graph $B$, as illustrated in Fig.~\ref{fig:localDecoder} left. Note that this construction has the same set of non-zero elements as the construction of Shi~\etal~\cite{SFJ14}, eqn.~(\ref{eqn:GT}), however the weights estimated by (\ref{eq:localDecoder}) will differ from $\G^{\top}$.

Memory vectors with indices $\Sset_0(i)$, contain other input vectors than $\x_i$. These vectors influence the estimate of $\q^{\top} \x_i$. Using the same reasoning as before, we propose to select indices $\Sset_1(i)$ of the memory units that contain any input vector $\x_j$ assigned to any memory unit with indices $\Sset_0(i)$. We call this selection 1-order local decoder, set $\Sset_1(i)$ contains indices of memory units connected by up to three edges to $\x_i$ in graph $B$, as illustrated in Fig.~\ref{fig:localDecoder} right.

The proposed selection of the non-zero entries of the decoder $\U$ efficiently picks relevant memory vectors to be used during the estimation of $\q^{\top} \x_i$ for each $\x_i$ in time complexity that is independent of the size of the input data. This construction is local, does not require the presence of all memory vectors in the memory and thus is suitable for batch processing and parallelization.

\subsubsection{Sparse decoder.}
\label{sec:sparseU}

To sparsify the decoder matrix $\U$ further,
we propose to add $\ell_1$-norm regularization into~\eqref{eq:localDecoder}:
\begin{equation}
\label{eq:sparseU}
\begin{aligned}
& \mathop{\text{minimize}}_{\u_i}
& & \frac{1}{2}\norm{\x_i - \Y_{\Sset(i)} \u_{i,\Sset(i)}}^2_2 + \lambda \| \u_{i,\Sset(i)} \|_1 
\end{aligned}
\end{equation}
In practice, we use a greedy algorithm called Orthogonal Matching Pursuit (OMP)~\cite{PCR+93,DMZ94}. 
OMP allows to choose the exact number of non-zero elements (instead of setting the parameter $\lambda$) in the solution so that  the complexity of the decoder $\U$ is directly adjusted. All the local properties of the decoder construction are preserved.

\subsubsection{Cascade decoder.}
\label{sec:addU}

The decoder $\U$ can be decomposed into two matrices $\U = \U^0 + \U^1$. Let $\c = \q^{\top} \Y$, the decoder is then written as:
\begin{equation} \label{eqn:decompo}
\q^{\top} \X \approx  \c \U^0 + \c \U^1\mbox{.}
\end{equation}
Since the estimated columns $\u_i$ of $\U$ contain a few significant elements (with high absolute value), some of them typically corresponding to $\Sset_0(i)$, and a larger number of less significant elements, the columns $\u^0_i$ of $\U^0$ are very sparse and contain the most significant entries, while the columns $\u^1_i$ of $\U^1$ contain the remaining elements of $\u_i$. The decomposition is then used in a cascade, first a short list of elements is efficiently obtained by using a rough approximation
\begin{equation}
\q^{\top} \X \approx  \hat{\s}^0 = \c \U^0 \mbox{.}
\end{equation}
Only for the shortlist $\Rset$, the finer estimate of the similarity (\ref{eqn:decompo}) is performed as 
$$
\q^{\top} \x_i \approx \hat{\s}^0 + \c \u^1_i \quad i \in \Rset \mbox{.}
$$
This process reduces the number of operations since $\U^0$ is much sparser than $\U$ and $|\Rset| \ll N$. The trade-off between the search accuracy and efficiency is controlled by the sparsity of $\U^0$. The memory requirements are increased only marginally, depending on the used representation of sparse matrices.

\subsubsection{Synergy with embedding methods.}
\label{sec:pq}
In this section, we discuss the combination of the proposed method with embedding methods, in particular product quantization (PQ)~\cite{JDS11}. In order to further reduce the memory footprint and the number of operations, the memory vectors $\y_i$ stored in $\y$ can be compressed by PQ. Approximating $\q^{\top} \y_i$ by asymmetric product quantizer is equivalent to actually evaluating $\q^{\top} \hat{\y}_i$, where $\hat{\y}_i$ is a quantized version of $\y_i$. Due to the local properties of the proposed algorithm, the decoder $\U$ can be construed from $\hat{\Y} = [\hat{\y}_1, \ldots, \hat{\y}_M]$ without any additional cost.

\subsection{Decoder correction}
\label{sec:corr}

The search results are affected by inaccuracy of the similarity $\hat{\s} \approx \q^{\top} \X$ estimate~\eqref{eq:decoding}.
The estimate $\hat{\s}$ can be used as a pre-filtering, true similarities can be computed for highly ranked vectors and these scores can be back-propagated to update other scores, as proposed by Shi~\etal~\cite{SFJ14}. 
This was shown to improve the search accuracy, but requires all dataset vectors to be accessible in the memory during query time. While such an approach reduces the complexity of the search (compared to the exhaustive search), it increases the memory footprint requirements.

Based on the orthogonal memory unit grouping, we propose a simple and efficient correction scheme, which 
 completely avoids the utilization of the input vectors and their true similarities.
Due to the orthogonality assignment, we make an assumption that there is at most one matching vector per memory unit to a query. In other words, if multiple vectors from the same memory unit score high in the ranked list, it is highly likely that only one is matching and the rest are false positives. We propose non-maxima suppression per memory unit. In practice, in a single pass through the ranked results, each top ranked (so far non-suppressed) result $\x_i$ suppresses all other vectors that appear in any of the memory vectors together with $\x_i$.   

Experimentally, we show that the orthogonality of the vectors is essential for the correction to work. 
This correction scheme improves the search accuracy, especially for memory units composed of smaller number $n$ of vectors.

\section{Experiments}
\label{sec:exp}
In this section, we experimentally verify the proposed method (all its components) in a large-scale image retrieval scenario. 
We first show the benefit of the orthogonal grouping and correction compared to random grouping. 
We define the following variants for our experiments.
\textbf{LO-GT} is memory vectors with orthogonal assignment (sec.~\ref{sec:omv}) and local decoder (sec.~\ref{sec:localDec}). \textbf{LO-GT*} additionally has the correction step (sec.~\ref{sec:corr}).
Their random counterparts are \textbf{RND} and \textbf{RND*}.
Memory vectors from memory units are created with the pseudo-inverse construction~\eqref{eq:pinv} for all cases.

\subsection{Experimental setup}

\head{Datasets.}
We use two well-known large-scale image retrieval benchmarks in our experiments: Oxford105k and Paris106k. 
They contain about 105k and 106k images respectively.
They are formed by adding 100k distractor images from Flickr~\cite{PCISZ07} to Oxford Buildings~\cite{PCISZ07} and Paris~\cite{PCISZ08} datasets.
We also perform a larger-scale experiment in revisited \roxf+\r1m~\cite{RIT+18}, which consists of new 1M challenging distractor set. We evaluate using the Medium setup.
Following the standard evaluation, the search performance is measured by mean average precision (mAP).

\head{Image Representation.}
We use state-of-the-art image descriptors extracted from a ResNet101 network fine-tuned for image retrieval~\cite{RTC17}.
Each descriptor is extracted from 3 different image scales using GeM pooling, and combined into a single descriptor as in~\cite{RIT+18}. Each descriptor has $d=2048$ dimensionality.

\head{Complexity Analysis.}
Following the existing work~\cite{IRF16}, efficiency is reported by measuring the complexity ratio.
This metric is based on the total number of scalar operations during the search.
It is computed as $\rho = (Md + s) / dN$, where $M$ is the total number of memory vectors, $d$ is the dimensionality, $s$ is the number of non-zero elements in the decoder $\U$, and $N$ is the number of images in the dataset.
Smaller $\rho$ means more efficient search.
Since our method does not require any image vectors to be loaded in the memory, $\rho$ also relates to the gain in memory footprint compared to the exhaustive search. The only exception is the cascade decoder, where some of the columns of $\U^1$ are not touched. 

\subsection{Retrieval performance}
We now compare our orthogonal assignment with random assignment in a retrieval scenario.
Various components of group testing, such as group size $n$, decoder order $l$ and sparsity of $\U$ are analyzed.
All the experiments in this subsection are performed on the Oxford105k dataset.

\begin{figure}
\centering
\input{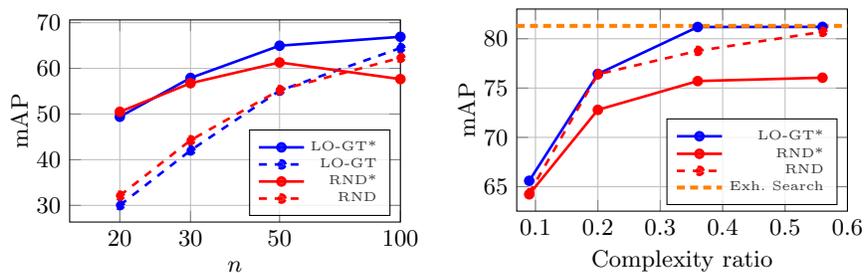}
\small
\begin{tabular}{cc}
%\extfig{birdsTrain}{
{
\begin{tikzpicture}
\begin{axis}[%
  width=0.49\linewidth,
  height=0.35\linewidth,
  xlabel={$n$},
  ylabel={mAP},
  legend cell align={left},
  legend pos=south east,
    legend style={cells={anchor=east}, font =\tiny, fill opacity=0.8, row sep=-2.5pt},
    xmax = 100,
    xmin = 15,
    xtick={20,30,50,100},
    xticklabels={20,30,50,100},
    grid=both,
   xmode = log,
    y label style={at={(axis description cs:-0.1,.5)}},
    x label style={at={(axis description cs:.5,-0.15)}}
]
  \addplot[color=blue,     solid, mark=*,  mark size=1.5, line width=1.0] table[x=n, y expr={100*\thisrow{omvc}}] \MmNn;\leg{LO-GT*};
  \addplot[color=blue,     dashed, mark=*,  mark size=1.5, line width=1.0] table[x=n, y expr={100*\thisrow{omv}}] \MmNn;\leg{LO-GT};
  \addplot[color=red,     solid, mark=*,  mark size=1.5, line width=1.0] table[x=n, y expr={100*\thisrow{randomc}}] \MmNn;\leg{RND*};
  \addplot[color=red,     dashed, mark=*,  mark size=1.5, line width=1.0] table[x=n, y expr={100*\thisrow{random}}] \MmNn;\leg{RND};

\end{axis}
\end{tikzpicture}
}

&

{
\begin{tikzpicture}
\begin{axis}[%
  width=0.49\linewidth,
  height=0.35\linewidth,
  xlabel={Complexity ratio},
  ylabel={mAP},
  legend cell align={left},
  legend pos=south east,
    legend style={cells={anchor=east}, font =\tiny, fill opacity=0.8, row sep=-2.5pt},
    xmax = 0.6,
    xmin = 0.07,
    xtick={0.1,0.2,0.3,0.4,0.5,0.6},
    xticklabels={0.1,0.2,0.3,0.4,0.5,0.6},
    grid=both,
   % xmode = log,
    y label style={at={(axis description cs:-0.1,.5)}},
    x label style={at={(axis description cs:.5,-0.15)}}
]
  \addplot[color=blue,     solid, mark=*,  mark size=1.5, line width=1.0] table[x=c, y expr={100*\thisrow{omvc}}] \firstOrd;\leg{LO-GT*};
  % \addplot[color=blue,     dashed, mark=*,  mark size=1.5, line width=1.0] table[x=c, y expr={100*\thisrow{omv}}] \firstOrd;\leg{OMV};  
  \addplot[color=red,     solid, mark=*,  mark size=1.5, line width=1.0] table[x=c, y expr={100*\thisrow{randomc}}] \firstOrd;\leg{RND*};
  \addplot[color=red,     dashed, mark=*,  mark size=1.5, line width=1.0] table[x=c, y expr={100*\thisrow{random}}] \firstOrd;\leg{RND};
  \addplot[color=orange, densely dashed, line width=1.5] coordinates {(0.07,81.3) (0.6,81.3)}; \leg{Exh. Search};
  % \addplot[color=red,     solid, mark=*,  mark size=1.5, line width=1.0] table[x=c, y expr={100*\thisrow{randomc}}] \zeroOrd;
  % \addplot[color=red,     dashed, mark=*,  mark size=1.5, line width=1.0] table[x=c, y expr={100*\thisrow{random}}] \zeroOrd;

  % \node [above] at (axis cs:  1.95,  81) {\scriptsize \textcolor{red}{$0.09$}};
  % \node [below] at (axis cs:  1.95,  21) {\scriptsize\textcolor{blue}{ $0.04$}};
  % \node [above] at (axis cs:  2.9,  83) {\scriptsize \textcolor{red}{$0.2$}};
  % \node [below] at (axis cs:  2.95,  50) {\scriptsize\textcolor{blue}{ $0.06$}};
  % \node [above] at (axis cs:  3.85,  83) {\scriptsize \textcolor{red}{$0.36$}};
  % \node [above] at (axis cs:  3.85,  61) {\scriptsize\textcolor{blue}{ $0.08$}};
  % \node [above] at (axis cs:  4.85,  83) {\scriptsize \textcolor{red}{$0.56$}};
  % \node [above] at (axis cs:  4.85,  63) {\scriptsize\textcolor{blue}{ $0.1$}};

\end{axis}
\end{tikzpicture}  

}

\end{tabular}
%\vspace{-10pt}
 \caption{\textbf{Left:} Impact of different $n$ for $m = n / 10$ and 0-order decoder on Oxford105k dataset. \textbf{Right: }Complexity ration vs.~retrieval quality for $l=1$-order decoding, $n = 50$, complexity ratio varies by changing $m = 2\ldots 5$.
 \label{fig:mnFig}
 }
\end{figure}

\head{Group size $n$.}
To keep the overall complexity fixed in this experiment, that is keeping sizes of $\Y$ and $\U$ constant, we set $m/n = M/N = 1 / 10$. Increasing the group size $n$ also increases the number $m$ of memory units each database vector is assigned to.
Figure~\ref{fig:mnFig} shows the mAP with the 0-order decoder for different values of $n$, comparing the proposed methods with the random grouping. For this settings, LO-GT without the decoder correction does not bring any significant improvement compared to random grouping. Using the decoder correction (LO-GT*) performs significantly better than the random grouping.
As shown in the same figure, the correction actually degrades the performance of random grouping after certain complexity.
This shows the benefit of our orthogonal grouping approach, which allows the correction process by assuming that there is only a single matching vector in the dataset.

\head{Performance of the 1-order decoder.} The 1-order decoder, non-sparse in this experiment, gives significantly better performance compared to 0-order decoder with higher complexity and memory footprint. 
In this experiment, we fix $n=50$ and show the mAP and complexity ($\rho$) for varying $m$ in Figure~\ref{fig:mnFig}. 
Our proposed method LO-GT* achieves the same mAP as exhaustive search with at the complexity ratio of $\rho = 0.36$ (corresponding to $m=4$), outperforming the random variants.

\begin{figure}
\centering
\input{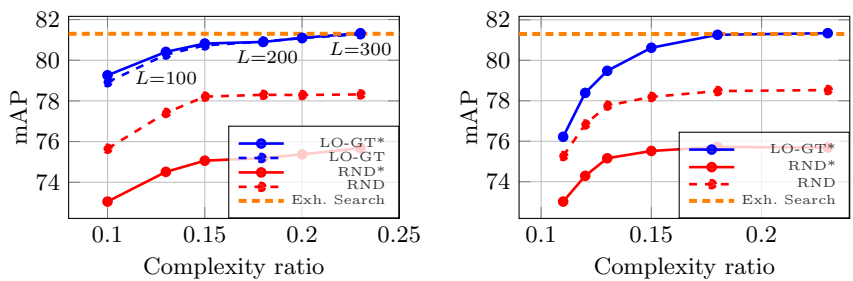}
\small
\begin{tabular}{cc}
%\extfig{birdsTrain}{
{
\begin{tikzpicture}
\begin{axis}[%
  width=0.49\linewidth,
  height=0.35\linewidth,
  xlabel={Complexity ratio},
  ylabel={mAP},
  legend cell align={right},
  legend pos=south east,
    legend style={at={(1,0)},cells={anchor=east}, font =\tiny, fill opacity=0.8, row sep=-3.5pt},
    xmax = 0.25,
    xmin = 0.08,
    %ymax = 84,
    % ymin = 60,
    % xmode = log,
    % xtick={50,100,250,500},
    % xticklabels={50 (\scriptsize{$0.1$}),100 (\scriptsize{$0.13$}),250 (\scriptsize{$0.2$}),$N$ (\scriptsize{$0.36$})},
    grid=both,
%    xmode = log,
    y label style={at={(axis description cs:-0.1,.5)}},
    x label style={at={(axis description cs:.5,-0.15)}}
]
  \addplot[color=blue,     solid, mark=*,  mark size=1.5, line width=1.0] table[x=c, y expr={100*\thisrow{omvc}}] \omp;\leg{LO-GT*};
  \addplot[color=blue,     dashed, mark=*,  mark size=1.5, line width=1.0] table[x=c, y expr={100*\thisrow{omv}}] \omp;\leg{LO-GT};
  \addplot[color=red,     solid, mark=*,  mark size=1.5, line width=1.0] table[x=c, y expr={100*\thisrow{randomc}}] \omp;\leg{RND*};
  \addplot[color=red,     dashed, mark=*,  mark size=1.5, line width=1.0] table[x=c, y expr={100*\thisrow{random}}] \omp;\leg{RND};  
  \addplot[color=orange, densely dashed, line width=1.5] coordinates {(0.08,81.3) (0.25,81.3)}; \leg{Exh. Search};

  % \node [below] at (axis cs:  0.097,  79.3) {\scriptsize \textcolor{black}{$L=50$}};
  % \node [below] at (axis cs:  0.127,  80.2) {\scriptsize \textcolor{black}{$L=100$}};
  % \node [below] at (axis cs:  0.152,  80.8) {\scriptsize \textcolor{black}{$L=150$}};
  % \node [below] at (axis cs:  0.182,  80.9) {\scriptsize \textcolor{black}{$L=200$}};
  % \node [below] at (axis cs:  0.202,  81.1) {\scriptsize \textcolor{black}{$L=250$}};
  % \node [below] at (axis cs:  0.23,  81.3) {\scriptsize \textcolor{black}{$L=300$}};

  %\node [above] at (axis cs:  0.10,  79) {\scriptsize \textcolor{black}{$L$=50}};
  \node [below] at (axis cs:  0.13,  79.9) {\scriptsize \textcolor{black}{$L$=100}};
  %\node [above] at (axis cs:  0.152,  81.5) {\scriptsize \textcolor{black}{$L$=150}};
  \node [below] at (axis cs:  0.182,  80.9) {\scriptsize \textcolor{black}{$L$=200}};
  %\node [above] at (axis cs:  0.202,  81.5) {\scriptsize \textcolor{black}{$L$=250}};
  \node [below] at (axis cs:  0.23,  81.3) {\scriptsize \textcolor{black}{$L$=300}};

\end{axis}
\end{tikzpicture}
}
&
{
  
\begin{tikzpicture}
\begin{axis}[%
  width=0.49\linewidth,
  height=0.35\linewidth,
  xlabel={Complexity ratio},
  ylabel={mAP},
  legend cell align={left},
  legend pos=south east,
    legend style={at={(1,0)},cells={anchor=east}, font =\tiny, fill opacity=0.8, row sep=-2.5pt},
    xmax = 0.24,
    xmin = 0.09,
    % ymax = 90,
    % ymin = 60,
    % xmode = log,
    % xtick={50,100,250,500},
    % xticklabels={50 (\scriptsize{$0.1$}),100 (\scriptsize{$0.13$}),250 (\scriptsize{$0.2$}),$N$ (\scriptsize{$0.36$})},
    grid=both,
%    xmode = log,
    y label style={at={(axis description cs:-0.1,.5)}},
    x label style={at={(axis description cs:.5,-0.15)}}
]
  \addplot[color=blue,     solid, mark=*,  mark size=1.5, line width=1.0] table[x=c, y expr={100*\thisrow{omvc}}] \additive;\leg{LO-GT*};
  \addplot[color=red,     solid, mark=*,  mark size=1.5, line width=1.0] table[x=c, y expr={100*\thisrow{randomc}}] \additive;\leg{RND*};
  \addplot[color=red,     dashed, mark=*,  mark size=1.5, line width=1.0] table[x=c, y expr={100*\thisrow{random}}] \additive;\leg{RND};  
  \addplot[color=orange, densely dashed, line width=1.5] coordinates {(0.09,81.3) (0.24,81.3)}; \leg{Exh. Search};

  % \node [below] at (axis cs:  0.097,  79.3) {\scriptsize \textcolor{black}{$L=50$}};
  % \node [below] at (axis cs:  0.127,  80.2) {\scriptsize \textcolor{black}{$L=100$}};
  % \node [below] at (axis cs:  0.152,  80.8) {\scriptsize \textcolor{black}{$L=150$}};
  % \node [below] at (axis cs:  0.182,  80.9) {\scriptsize \textcolor{black}{$L=200$}};
  % \node [below] at (axis cs:  0.202,  81.1) {\scriptsize \textcolor{black}{$L=250$}};
  % \node [below] at (axis cs:  0.23,  81.3) {\scriptsize \textcolor{black}{$L=300$}};

  % \node [below] at (axis cs:  50,  77) {\scriptsize\textcolor{blue}{ $0.1$}};
  % \node [below] at (axis cs:  100,  75) {\scriptsize \textcolor{red}{$0.13$}};
  % \node [below] at (axis cs:  100,  80) {\scriptsize\textcolor{blue}{ $0.04$}};
  % \node [above] at (axis cs:  200,  79) {\scriptsize \textcolor{red}{$0.09$}};
  % \node [below] at (axis cs:  200,  82) {\scriptsize\textcolor{blue}{ $0.04$}};
  % \node [above] at (axis cs:  250,  82) {\scriptsize \textcolor{red}{$0.09$}};
  % \node [below] at (axis cs:  250,  80) {\scriptsize\textcolor{blue}{ $0.04$}};
  % \node [above] at (axis cs:  500,  83) {\scriptsize \textcolor{red}{$0.09$}};
  % \node [below] at (axis cs:  500,  81) {\scriptsize\textcolor{blue}{ $0.04$}};

\end{axis}
\end{tikzpicture}

}

\end{tabular}
%\vspace{-10pt}
 \caption{\textbf{Left:} Search accuracy for sparse $\U$ of 1-order decoder. $L$ is the number of non-zeros in each column of $\U$. \textbf{Right: }Impact of cascade decoder. Complexity ratio is varied by changing $p = 50\% \ldots 100\%$. $\U^0$ is populated with the most significant entries carrying $p$ percent of the column energy. Rest of non-zero entries are assigned to $\U_1$.
 \label{fig:fig_omp}
 }
\end{figure}

\head{Sparse decoder}
is obtained by adding a sparsity constraint on the solution of $\U$~\eqref{eq:sparseU}.
We show the impact of such sparsity constraint in Figure~\ref{fig:fig_omp}.
Orthogonal Matching Pursuit algorithm is used to adjust the number of non-zeros ($L$) on each column of $\U$.
Smaller values $L$ leads to a sparser solution, hence lower complexity ratio.
It is shown that we achieve better accuracy than the random variants for all values of $L$. 
Setting $L=300$ gives us the same accuracy as the exhaustive search.
This corresponds to the complexity ratio of $0.23$.
As the complexity of the decoder increases, the benefit of the correction is less pronounced. At the same time
LO-GT without correction, \ie due to better estimates of the similarity, significantly outperforms both random grouping variants.

\head{Cascade decoder} is the decomposition of the decoder $\U$ in two matrices $\U = \U^0 + \U^1$~(sec.~\ref{sec:addU}).
In this experiment, $n=50$, $m=4$ and the sparse decoder $\U$ with $L=300$ from the previous experiment is decomposed it into two matrices. Each column of $\U^0$ is populated with the most significant entries carrying $p$ percent of the column energy.
Remaining non-zero entries are assigned to $\U_1$. Figure~\ref{fig:fig_omp} shows the outcome of this approach.
Different complexity ratio is obtained by changing $p = 50\% \ldots 100\%$. It is shown that the cascade decoder reduces the complexity ratio even further without harming the search accuracy.
\head{Actual search time} is measured as seconds instead of complexity ratio.  In a single-thread CPU, exhaustive search in Oxford105k dataset takes 0.198~$s$ per query on average. 
Our search time (when the complexity ratio is 0.18) in the same environment is 0.054 $s$ per query. 
That's a ratio of 0.27, but we would like to note that we use a simple Matlab implementation which is not optimized for this task.

\subsection{Comparison with other methods}

We compare our method against the existing group testing techniques in the literature~\cite{SFJ14,IRF16}.
Two of our comparisons are made against the two variants of the group testing framework proposed by Shi~\etal~\cite{SFJ14}. 
The first variant involves computing the full pseudo-inverse of the assignment graph $\G$. 
Since the resulting matrix is a large dense matrix, this variant is not efficient and is only included to serve as a baseline. 
We also compare against the back-propagation variant proposed by Shi~\etal~\cite{SFJ14}, where the scores are updated based on the true distance computations with dataset vectors. 
This variant is more efficient but requires higher memory footprint than the exhaustive search.
All dataset vectors, in addition to memory vectors, need to be available in the memory.

We also compare against the dictionary learning solution proposed by Iscen~\etal~\cite{IRF16}. 
This method also has two variants. 
The first case involves using the entire the dataset to learn group vectors.
Offline processing of this variant is not efficient for practical large-scale applications. 
Optimization problem~\eqref{eq:dictLearn} takes a long time to learn the group vectors.
Alternatively, coresets~\cite{AHV04,FFS13} are proposed to reduce the indexing time. 
Coresets are representative data points sampled from the dataset. 
Number of coresets is set to $N/5$. 

\begin{table}
\centering{\scriptsize
    \begin{tabular}{|@{\msp}l@{\msp}@{\msp}c@{\msp}@{\msp}c@{\msp}@{\msp}r@{\msp}@{\msp}r@{\msp}@{\msp}r@{\msp}@{\msp}r@{\msp}|}
    \hline                          
    \multicolumn{3}{|@{\msp}c@{\msp}}{} &  \multicolumn{2}{c@{\msp}}{\textbf{Oxford105k}}  &   \multicolumn{2}{@{\msp}c@{\msp}|}{\textbf{Paris106k}}  \\
                                    & Complexity  		    & Memory            & Index Time	& mAP               & Index Time      & mAP          \\ \hline
     Baseline                       & 1.00\                 & 1.00\             & -             & 81.3\             & -               & 83.4\        \\ \hline \hline
     GT~\cite{SFJ14} pinv    		& 4.20\                 & 4.20\             & $<$1\         & 63.9\             & $<$1\           & 56.5\        \\
     GT~\cite{SFJ14} w/ bp.  		& 0.36\                 & 1.36\             & $<$1\         & 73.4\             & $<$1\           & 73.6\        \\
     DL~\cite{IRF16} w/ cset        & 0.18\                 & 0.18\             & 273\          & 81.4\             & 288\            & 85.2\        \\ 
     DL~\cite{IRF16}                & 0.11\                 & 0.11\             & 435\          & 86.8\             & 492\            & 86.2\        \\ \hline \hline
     Ours, $l=1$                    & 0.18\                 & 0.23\             & 4\            & 81.3\             & 4\              & 83.7\        \\ \hline

\end{tabular}}

\caption{Comparison of our method and existing group testing methods. Index time (in minutes) is the time it takes to create group vectors and the decoder. We set $M=8408$ for all methods. We set the size of coresets $N/5$ for DL~\cite{IRF16}.
\label{tab:soaGt}
}
\end{table}

Table~\ref{tab:soaGt} shows the comparison between our method and prior art in group testing algorithms.
Timings are reported on a server with 32 cores.
For every variant we set $M =8408$, which corresponds to $n=50$ and $m=4$ for~\cite{SFJ14} and the proposed method.
Note that DL has higher mAP but this comes at a significant offline cost.
Finding group vectors with DL~\cite{IRF16} involves solving an expensive optimization problem.
It also requires entire dataset to be loaded in the memory at once, limiting its scalability for very large $N$.
Furthermore, it is shown that the offline complexity of DL increases exponentially as $N$ increases~\cite{IRF16}.
Therefore, this method is scalable only if the size of the dataset is reduced with coresets.
Search accuracy is lower in that case, but the offline indexing time is still significantly higher than our method.

\head{Batch processing.} 
We evaluate the search performance in a scenario where the data becomes available in batches over time.
We divide the Oxford1M dataset randomly into $b$ batches of equal size and create memory vectors $\Y_b$ and the decoding matrix $\U_b$ separately for each batch.
After processing all the batches, we concatenate all $\Y_b$ and $\U_b$ and perform the search.
Figure~\ref{fig:fig_stream} shows the mAP for different number of batches. 
It can be observed that the search accuracy of DL~\cite{IRF16} degrades significantly as the data are divided in more batches.
This can be explained by the nature of this method, the fewer data are used in the global optimization, the less efficient search.
The indexing time of DL is extremely high. Indexing the whole dataset in one go is not tractable, and for $b = 10$ it takes about 50 hours. On the contrary, our method can handle any scenario where matching vectors become available over time.
This clearly showns in Figure~\ref{fig:fig_stream}, where the performance of our method is stable regardless of $b$. 

\begin{figure}
\centering
\input{figs/data/sample}
\small
\begin{tabular}{cc}
{
%\extfig{birdsTrain}{
\begin{tikzpicture}
\begin{axis}[%
  width=0.49\linewidth,
  height=0.35\linewidth,
  xlabel={Number of batches $b$},
  ylabel={mAP},
  legend cell align={left},
  legend pos=south west,
    legend style={cells={anchor=west}, font =\tiny, fill opacity=0.8, row sep=-2.5pt},
    xmax = 110,
    xmin = 1,
    % ymax = 90,
    % ymin = 60,
    % xmode = log,
    xtick={10,20,50,100},
    grid=both,
   % xmode = log,
    y label style={at={(axis description cs:-0.1,.5)}},
    x label style={at={(axis description cs:.5,-0.15)}}
]
  \addplot[color=orange, densely dashed, line width=1.5] coordinates {(1,41.5) (110,41.5)}; \leg{Exh. Search};
  \addplot[color=blue,     solid, mark=*,  mark size=1.5, line width=1.0] 
  plot [error bars/.cd, y dir = both, y explicit, error bar style={line width=1pt,solid},error mark options={line width=.5pt,mark size=1.5pt,rotate=90}]
  table[x=b, y expr={100*\thisrow{omvc}}, y error = errors] \rstream;\leg{Ours};
  \addplot[color=red,     solid, mark=*,  mark size=1.5, line width=1.0] table[x=b, y expr={100*\thisrow{dl}}] \rstream;\leg{DL~\cite{IRF16}};

\end{axis}
\end{tikzpicture}
}
&
{
\begin{tikzpicture}
\begin{axis}[%
  width=0.49\linewidth,
  height=0.35\linewidth,
  xlabel={Complexity ratio},
  ylabel={mAP},
  legend cell align={left},
  legend pos=south east,
    legend style={cells={anchor=west}, font =\tiny, fill opacity=0.8, row sep=-2.5pt},
    xmax = 0.8,
    xmin = 0.01,
    % ymax = 90,
    % ymin = 60,
    % xmode = log,
    % xtick={50,100,250,500},
    % xticklabels={50 (\scriptsize{$0.1$}),100 (\scriptsize{$0.13$}),250 (\scriptsize{$0.2$}),$N$ (\scriptsize{$0.36$})},
    grid=both,
%    xmode = log,
    y label style={at={(axis description cs:-0.1,.5)}},
    x label style={at={(axis description cs:.5,-0.15)}}
]
  \addplot[color=blue,     solid, mark=*,  mark size=1.5, line width=1.0] table[x=c, y expr={100*\thisrow{omvc}}] \gtVsGraph;\leg{LO-GT*-$p$};
  \addplot[color=cyan,     solid, mark=*,  mark size=1.5, line width=1.0] table[x=c, y expr={100*\thisrow{omvcS}}] \gtVsGraph;\leg{LO-GT*-$p$ ($s$)};
  \addplot[color=black!40!green,     solid, mark=o,  mark size=1.5, line width=1.0] table[x=c, y expr={100*\thisrow{omvcOMP}}] \gtVsGraph;\leg{LO-GT*-$L$};
  \addplot[color=green,     solid, mark=o,  mark size=1.5, line width=1.0] table[x=c, y expr={100*\thisrow{omvcOMPS}}] \gtVsGraph;\leg{LO-GT*-$L$ ($s$)};
  \addplot[color=red,     solid, mark=*,  mark size=1.5, line width=1.0] table[x=c, y expr={100*\thisrow{hnsw}}] \gtVsGraph;\leg{HNSW ($s$)};
  \addplot[color=orange, densely dashed, line width=1.5] coordinates {(0.01,81.3) (0.8,81.3)}; \leg{Exh. Search};

  % \node [below] at (axis cs:  0.097,  79.3) {\scriptsize \textcolor{black}{$L=50$}};
  % \node [below] at (axis cs:  0.127,  80.2) {\scriptsize \textcolor{black}{$L=100$}};
  % \node [below] at (axis cs:  0.152,  80.8) {\scriptsize \textcolor{black}{$L=150$}};
  % \node [below] at (axis cs:  0.182,  80.9) {\scriptsize \textcolor{black}{$L=200$}};
  % \node [below] at (axis cs:  0.202,  81.1) {\scriptsize \textcolor{black}{$L=250$}};
  % \node [below] at (axis cs:  0.23,  81.3) {\scriptsize \textcolor{black}{$L=300$}};

  % \node [below] at (axis cs:  50,  77) {\scriptsize\textcolor{blue}{ $0.1$}};
  % \node [below] at (axis cs:  100,  75) {\scriptsize \textcolor{red}{$0.13$}};
  % \node [below] at (axis cs:  100,  80) {\scriptsize\textcolor{blue}{ $0.04$}};
  % \node [above] at (axis cs:  200,  79) {\scriptsize \textcolor{red}{$0.09$}};
  % \node [below] at (axis cs:  200,  82) {\scriptsize\textcolor{blue}{ $0.04$}};
  % \node [above] at (axis cs:  250,  82) {\scriptsize \textcolor{red}{$0.09$}};
  % \node [below] at (axis cs:  250,  80) {\scriptsize\textcolor{blue}{ $0.04$}};
  % \node [above] at (axis cs:  500,  83) {\scriptsize \textcolor{red}{$0.09$}};
  % \node [below] at (axis cs:  500,  81) {\scriptsize\textcolor{blue}{ $0.04$}};

\end{axis}
\end{tikzpicture}

}

\end{tabular}
%\vspace{-10pt}
 \caption{\textbf{Left:} The mAP of the indexing structure constructed sequentially. \roxf+\r1m is divided into $b$ batches and $\Y_b$ and $\U_b$ is created separately for each batch. We run our method multiple times and show the standard variation in vertical bars. \textbf{Right:} Comparison between our methods and HNSW~\cite{JDH17} in Oxford105k. Curves with the complexity ratio based on measured run-time are denoted by $(s)$, otherwise measured by the number of scalar operations.
 \label{fig:fig_stream}
 }
\end{figure}

\head{Comparison with partitioning.}  Finally, we compare our framework against the well-known FLANN toolbox~\cite{ML14}.
We use the ``autotuned'' setting of FLANN, setting the target precision to 0.95. 
The average speed-up ratio after 5 runs is 1.45, which corresponds to a complexity ratio of 0.69.
Compared to that, our method is able to achieve the baseline mAP performance with only 0.2 complexity ratio and memory footprint.

\head{Comparison with graph-based indexing.}
Finally, we compare our method against a popular graph-based method by Malkov and Yashunin~\cite{JDH17}. 
We use the implementation provided in FAISS toolbox~\cite{MY16}. 
To report the complexity ratio, a ratio of approximate-search time using the HNSW index to an exhaustive-search time, both on a single-thread CPU, is computed. 
A comparison of mAP on Oxford105k for different complexity ratios is shown in Fig~\ref{fig:fig_stream}(right plot). 
Two variants of the proposed method are shown: sparse 1-order decoder and cascade decoder $\mathbf{U}_0 + \mathbf{U}_1$ (as in Fig.~\ref{fig:fig_omp}), where the complexity is controlled by parameter $L$ and $p$ respectively. 
Additionally, we also report complexity ratio measured as a ratio of approximate-search and exhaustive-search times (as in HNSW). 
The plot shows that our method clearly outperforms HNSW. Furthermore, our framework requires significantly smaller memory footprint. 

\head{Combination with PQ. }
As described in Section~\ref{sec:pq}, our method is compatible with existing embedding techniques, such as product quantization (PQ)~\cite{JDS11}.
We compress 2048D memory vectors into $c$ bytes.
Table~\ref{tab:pq} shows the mAP for different $c$.
Note that learning $\U$ from compressed $\hat{\Y}$ (denoted by $\U^{\text{pq}}$ in the table) is important in this case.
If $\U$ is learned from non-compressed $\Y$, then mAP is significantly degraded.

\begin{SCtable}
\centering{\scriptsize
    \begin{tabular}{|@{\sssp}l@{\sssp}|@{\sssp}c@{\sssp}@{\sssp}c@{\sssp}@{\sssp}c@{\sssp}|}
    \hline                          
     \backslashbox{Decoder}{$c$}    & 256         & 32        & 16           \\ \hline
     $\U$ from $\Y$    		                & 77.7\     & 52.2\     & 44.8\         \\
     $\U^{\text{pq}}$ from $\hat\Y$  	            & 79.2\     & 70.5\     & 62.3\         \\ \hline
\end{tabular}}

\caption{Combination of PQ and our method. Each memory vector is compressed into $c$ bytes.
\label{tab:pq}
}
\end{SCtable}

\section{Conclusions}
\label{sec:conc}
We have proposed two contributions to the group testing framework. First, the linear-time complexity orthogonal grouping increases the probability that at most one element from each group is matching to a given query. Non-maxima suppression with each group efficiently reduces the number of false positive results at no extra cost. Second, unlike in other similarly performing approaches, such as dictionary learning~\cite{IRF16}, all processing is local, orders of magnitude faster, and suitable to process data in batches and in parallel.
%The method offers a trade-off between the search accuracy and search and memory efficiency. 
We experimentally show, that for any choice of the efficiency, the proposed method significantly outperforms previously used random grouping. 
Finally, the proposed method achieves search accuracy of the exhaustive search with significant reduction in the search complexity.
%The method is naturally combined with existing embedding methods, which further increases its efficiency.

%\scriptsize{
\head{Acknowledgments}
The authors were supported by MSMT LL1303 ERC-CZ grant and the OP VVV funded project 
CZ.02.1.01/0.0/0.0/16\_019/0000765 
``Research Center for Informatics''.
%}

\bibliographystyle{splncs}
\bibliography{egbib}

\begin{thebibliography}{10}

\bibitem{SZ03}
Sivic, J., Zisserman, A.:
\newblock {Video Google}: {A} text retrieval approach to object matching in
  videos.
\newblock In: ICCV. (2003)

\bibitem{PCISZ07}
Philbin, J., Chum, O., Isard, M., Sivic, J., Zisserman, A.:
\newblock Object retrieval with large vocabularies and fast spatial matching.
\newblock In: CVPR. (June 2007)

\bibitem{PCISZ08}
Philbin, J., Chum, O., Isard, M., Sivic, J., Zisserman, A.:
\newblock Lost in quantization: Improving particular object retrieval in large
  scale image databases.
\newblock In: CVPR. (June 2008)

\bibitem{JDS10a}
J\'egou, H., Douze, M., Schmid, C.:
\newblock Improving bag-of-features for large scale image search.
\newblock IJCV \textbf{87}(3) (February 2010)  316--336

\bibitem{JSHV10}
J\'egou, H., Schmid, C., Harzallah, H., Verbeek, J.:
\newblock Accurate image search using the contextual dissimilarity measure.
\newblock IEEE Trans. PAMI \textbf{32}(1) (January 2010)  2--11

\bibitem{GVSG10}
van Gemert, J.C., Veenman, C., Smeulders, A.W., Geusebroek, J.:
\newblock Visual word ambiguity.
\newblock IEEE Trans. PAMI \textbf{32}(7) (July 2010)  1271--1283

\bibitem{BL12}
Babenko, A., Lempitsky, V.:
\newblock The inverted multi-index.
\newblock In: CVPR. (June 2012)

\bibitem{JDSP10}
J\'egou, H., Douze, M., Schmid, C., P\'erez, P.:
\newblock Aggregating local descriptors into a compact image representation.
\newblock In: CVPR. (June 2010)

\bibitem{PD07}
Perronnin, F., Dance, C.R.:
\newblock Fisher kernels on visual vocabularies for image categorization.
\newblock In: CVPR. (June 2007)

\bibitem{PSM10}
Perronnin, F., J.S\'anchez, Mensink, T.:
\newblock Improving the fisher kernel for large-scale image classification.
\newblock In: ECCV. (September 2010)

\bibitem{BSCL14}
Babenko, A., Slesarev, A., Chigorin, A., Lempitsky, V.:
\newblock Neural codes for image retrieval.
\newblock In: ECCV. (2014)

\bibitem{TSJ15}
Tolias, G., Sicre, R., J{\'e}gou, H.:
\newblock Particular object retrieval with integral max-pooling of cnn
  activations.
\newblock ICLR (2016)

\bibitem{GARL16}
Gordo, A., Almazan, J., Revaud, J., Larlus, D.:
\newblock Deep image retrieval: Learning global representations for image
  search.
\newblock ECCV (2016)

\bibitem{RTC16}
Radenovi{\'c}, F., Tolias, G., Chum, O.:
\newblock {CNN} image retrieval learns from bow: Unsupervised fine-tuning with
  hard examples.
\newblock ECCV (2016)

\bibitem{RTC17}
Radenovi{\'c}, F., Tolias, G., Chum, O.:
\newblock Fine-tuning cnn image retrieval with no human annotation.
\newblock arXiv preprint arXiv:1711.02512 (2017)

\bibitem{ML14}
Muja, M., Lowe, D.G.:
\newblock Scalable nearest neighbor algorithms for high dimensional data.
\newblock IEEE Trans. PAMI \textbf{36} (2014)

\bibitem{NS06}
Nist\'er, D., Stew\'enius, H.:
\newblock Scalable recognition with a vocabulary tree.
\newblock In: CVPR. (June 2006)  2161--2168

\bibitem{IM98}
Indyk, P., Motwani, R.:
\newblock Approximate nearest neighbors: towards removing the curse of
  dimensionality.
\newblock In: STOC. (1998)  604--613

\bibitem{GIM99}
Gionis, A., Indyk, P., Motwani, R.:
\newblock Similarity search in high dimension via hashing.
\newblock In: VLDB. (1999)  518--529

\bibitem{DIIM04}
Datar, M., Immorlica, N., Indyk, P., Mirrokni, V.:
\newblock Locality-sensitive hashing scheme based on p-stable distributions.
\newblock In: Proceedings of the Symposium on Computational Geometry. (2004)

\bibitem{LCL04}
Lv, Q., Charikar, M., Li, K.:
\newblock Image similarity search with compact data structures.
\newblock In: CIKM. (November 2004)  208--217

\bibitem{NPF12}
Norouzi, M., Punjani, A., Fleet, D.J.:
\newblock Fast search in hamming space with multi-index hashing.
\newblock In: CVPR. (2012)

\bibitem{WTF09}
Weiss, Y., Torralba, A., Fergus, R.:
\newblock Spectral hashing.
\newblock In: NIPS. (December 2009)

\bibitem{RL10}
Raginsky, M., Lazebnik, S.:
\newblock Locality-sensitive binary codes from shift-invariant kernels.
\newblock In: NIPS. (2010)

\bibitem{JDS11}
J\'egou, H., Douze, M., Schmid, C.:
\newblock Product quantization for nearest neighbor search.
\newblock IEEE Trans. PAMI \textbf{33}(1) (January 2011)  117--128

\bibitem{GHKS13}
Ge, T., He, K., Ke, Q., Sun, J.:
\newblock Optimized product quantization for approximate nearest neighbor
  search.
\newblock In: CVPR. (June 2013)

\bibitem{KA14}
Kalantidis, Y., Avrithis, Y.:
\newblock Locally optimized product quantization for approximate nearest
  neighbor search.
\newblock In: CVPR. (2014)

\bibitem{JZPG17}
Jain, H., Zepeda, J., P{\'e}rez, P., Gribonval, R.:
\newblock Subic: A supervised, structured binary code for image search.
\newblock In: ICCV. (2017)

\bibitem{JZPG18}
Jain, H., Zepeda, J., P{\'e}rez, P., Gribonval, R.:
\newblock Learning a complete image indexing pipeline.
\newblock (2018)

\bibitem{D43}
Dorfman, R.:
\newblock The detection of defective members of large populations.
\newblock The Annals of Mathematical Statistics \textbf{14}(4) (1943)  436--440

\bibitem{IFGRJ14}
Iscen, A., Furon, T., Gripon, V., Rabbat, M., J{\'e}gou, H.:
\newblock Memory vectors for similarity search in high-dimensional spaces.
\newblock IEEE Trans. Big Data \textbf{4}(1) (2018)

\bibitem{SFJ14}
Shi, M., Furon, T., J\'egou, H.:
\newblock A group testing framework for similarity search in high-dimensional
  spaces.
\newblock In: ACM Multimedia. (November 2014)

\bibitem{IAF16}
Iscen, A., Amsaleg, L., Furon, T.:
\newblock Scaling group testing similarity search.
\newblock In: ACM ICMR. (2016)

\bibitem{IRF16}
Iscen, A., Rabbat, M., Furon, T.:
\newblock Efficient large-scale similarity search using matrix factorization.
\newblock In: CVPR. (2016)

\bibitem{RM71}
Rao, C.R., Mitra, S.K.:
\newblock Generalized inverse of matrices and its applications. Volume~7.
\newblock (1971)

\bibitem{ABJ14}
Aldridge, M., Baldassini, L., Johnson, O.:
\newblock Group testing algorithms: bounds and simulations.
\newblock IEEE Trans. Inform. Theory (2014)

\bibitem{BBIAD11}
Bickson, D., Baron, D., Ihler, A., Avissar, H., Dolev, D.:
\newblock Fault identification via nonparametric belief propagation.
\newblock {\sc IEEE} Transactions on Signal Processing \textbf{59}(6) (2011)
  2602--2613

\bibitem{CHKV09}
Cheraghchi, M., Hormati, A., Karbasi, A., Vetterli, M.:
\newblock Compressed sensing with probabilistic measurements: A group testing
  solution.
\newblock In: 47th Annual Allerton Conference on Communication, Control, and
  Computing. (2009)

\bibitem{GI10}
Gilbert, A., Indyk, P.:
\newblock Sparse recovery using sparse matrices.
\newblock Proceedings of the IEEE \textbf{98}(6) (2010)  937--947

\bibitem{SeJ10}
Sejdinovic, D., Johnson, O.:
\newblock Note on noisy group testing: asymptotic bounds and belief propagation
  reconstruction.
\newblock In: 48th Annual Allerton Conference on Communication, Control, and
  Computing. (2010)

\bibitem{PCR+93}
Pati, Y.C., Rezaiifar, R., Krishnaprasad, P.:
\newblock Orthogonal matching pursuit: Recursive function approximation with
  applications to wavelet decomposition.
\newblock In: ASILOMAR. (1993)  40--44

\bibitem{DMZ94}
Davis, G.M., Mallat, S.G., Zhang, Z.:
\newblock Adaptive time-frequency decompositions with matching pursuit.
\newblock In: SPIE's International Symposium on Optical Engineering and
  Photonics in Aerospace Sensing. (1994)  402--413

\bibitem{RIT+18}
Radenovi{\'c}, F., Iscen, A., Tolias, G., Avrithis, Y., Chum, O.:
\newblock Revisiting oxford and paris: Large-scale image retrieval
  benchmarking.
\newblock In: CVPR. (2018)

\bibitem{AHV04}
Agarwal, P.K., Har-Peled, S., Varadarajan, K.R.:
\newblock Approximating extent measures of points.
\newblock Journal of the ACM \textbf{51}(4) (2004)  606--635

\bibitem{FFS13}
Feldman, D., Feigin, M., Sochen, N.:
\newblock Learning big (image) data via coresets for dictionaries.
\newblock Journal of Mathematical Imaging and Vision \textbf{46}(3) (2013)
  276--291

\bibitem{JDH17}
Johnson, J., Douze, M., J{\'e}gou, H.:
\newblock Billion-scale similarity search with gpus.
\newblock arXiv preprint arXiv:1702.08734 (2017)

\bibitem{MY16}
Malkov, Y.A., Yashunin, D.:
\newblock Efficient and robust approximate nearest neighbor search using
  hierarchical navigable small world graphs.
\newblock arXiv preprint arXiv:1603.09320 (2016)

\end{thebibliography}
\end{document}